\documentclass[lettersize,journal]{IEEEtran}

\usepackage{amsmath,amssymb,amsfonts}
\usepackage{mathtools}
\usepackage{graphicx}
\usepackage{hyperref}
\usepackage{cleveref}
\usepackage{url}
\usepackage[bottom]{footmisc}
\usepackage{cite}
\usepackage{orcidlink}
\usepackage{algorithm}
\usepackage{algorithmicx}
\usepackage{algpseudocode}

\usepackage{amsthm}

\theoremstyle{definition}
\newtheorem{definition}{Definition}

\crefname{figure}{Fig.}{Fig.}

\title{DiAReL: Reinforcement Learning with Disturbance Awareness for Robust Sim2Real Policy Transfer in Robot Control}

\author{Mohammadhossein Malmir\,\orcidlink{0000-0003-0610-7899},~\IEEEmembership{Member,~IEEE,} %
Josip Josifovski\,\orcidlink{0000-0002-1031-7621},~\IEEEmembership{Member,~IEEE,} \\%
Noah Klarmann\,\orcidlink{0009-0008-9157-9228}, %
and Alois Knoll\,\orcidlink{0000-0003-4840-076X},~\IEEEmembership{Fellow,~IEEE}
\thanks{This work was supported by the research project "A-IQ READY" funded within the Chips Joint Undertaking (Chips JU) - the Public-Private Partnership for research, development, and innovation under Horizon Europe – and National Authorities under grant agreement No. 101096658. (Corresponding Author: Mohammadhossein Malmir.)}
\thanks{This paper has supplementary downloadable material, provided by the authors. This includes a multimedia MP4 format movie clip, which shows the sim2real experiment runs. This material is 4.93 MB in size.}%
\thanks{Mohammadhossein Malmir, Josip Josifovski, and Alois Knoll are with the Department of Computer Engineering, School of Computation, Information and Technology, Technical University of Munich, Munich, Germany (email: hossein.malmir@tum.de; josip.josifovski@tum.de; k@tum.de).}%
\thanks{Noah Klarmann is with the Rosenheim University of Applied Sciences, Rosenheim, Germany (email: noah.klarmann@th-rosenheim.de).}%
\vspace{-1.5\baselineskip}
}

\hypersetup{
  pdftitle={DiAReL: Reinforcement Learning with Disturbance Awareness for Robust Sim2Real Policy Transfer in Robot Control},
  pdfauthor={Mohammadhossein Malmir, Josip Josifovski, Noah Klarmann, and Alois Knoll},
  pdfsubject={Robotics (cs.RO); Machine Learning (cs.LG); Systems and Control (cs.SY,eess.SY)}
  pdfkeywords={Reinforcement Learning, Robust Control, Markov Decision Processes, Disturbance Observers, Sim2Real Transfer},%
  colorlinks=true,
  linktoc=all,
  linkbordercolor={1 1 1},
  linkcolor=black,
  citecolor=black,
  urlcolor=black,
}

\begin{document}

\maketitle
\thispagestyle{empty}
\pagestyle{empty}

\begin{abstract}

Delayed Markov decision processes (DMDPs) fulfill the Markov property by augmenting the state space of agents with a finite time window of recently committed actions. In reliance on these state augmentations, delay-resolved reinforcement learning algorithms train policies to learn optimal interactions with environments featuring observation or action delays. Although such methods can be directly trained on the real robots, due to sample inefficiency, limited resources, or safety constraints, a common approach is to transfer models trained in simulation to the physical robot. However, robotic simulations rely on approximated models of the physical systems, which hinders the sim2real transfer. In this work, we consider various uncertainties in modeling the robot or environment dynamics as unknown intrinsic disturbances applied to the system input. We introduce the disturbance-augmented Markov decision process (DAMDP) in delayed settings as a novel representation to incorporate disturbance estimation in training on-policy reinforcement learning algorithms. The proposed method is validated across several metrics on learning robotic reaching and pushing tasks and compared with disturbance-unaware baselines. The results show that the disturbance-augmented models can achieve higher stabilization and robustness in the control response, which in turn improves the prospects of successful sim2real transfer.

\end{abstract}


\section{INTRODUCTION}

\IEEEPARstart{D}{espite} the recent advances in devising sample-efficient reinforcement learning (RL) algorithms like model-based or offline RL, there are still many limitations for direct training of the RL agents on actual robots \cite{Peng.2018} and only a few successes \cite{Levine.07Mar16,Levine.03Apr15}. Comparatively, state-of-the-art model-free RL algorithms have primarily achieved astonishing results when trained by abundant low-cost synthetic data from simulations. Hence, a common approach in robot learning is to conduct agent training in a simulated environment that closely mimics the real world \cite{Peng.2018,Tobin.20Mar17,Tan.27Apr18,Christiano.11Oct16,Oliva2022,AndreiA.Rusu.2017}.

Conventional RL algorithms usually assume a consistent environment for both the training and testing phases \cite{Tessler.26Jan19}, which makes them unable to generalize to slightly varied dynamics in the environment. As a result, these algorithms can experience a decline in performance when transferred to the real world \cite{Lillicrap.10Sep15} due to the existing \textit{reality gap} \cite{Tobin.20Mar17,Tan.27Apr18}.

The first step towards closing the gap between simulation and the real world is to improve the simulation's accuracy in both the physics and the perception aspects \cite{Peng.2018,Tan.27Apr18,Zhu.24Oct17}. However, since there are always differences between simulation and the real world, techniques like domain adaptation \cite{Tzeng.23Nov15,Bousmalis.22Sep17} and continual learning \cite{AndreiA.Rusu.2017,Josifovski2020Continual,Josifovski2024} are contrived to allow the RL agent to adapt its behavior by continuing the learning process in the real world as well \cite{AndreiA.Rusu.2017,Tremblay.18Apr18}.

\begin{figure}[t]%
	\centering%
	\includegraphics[width=\columnwidth]{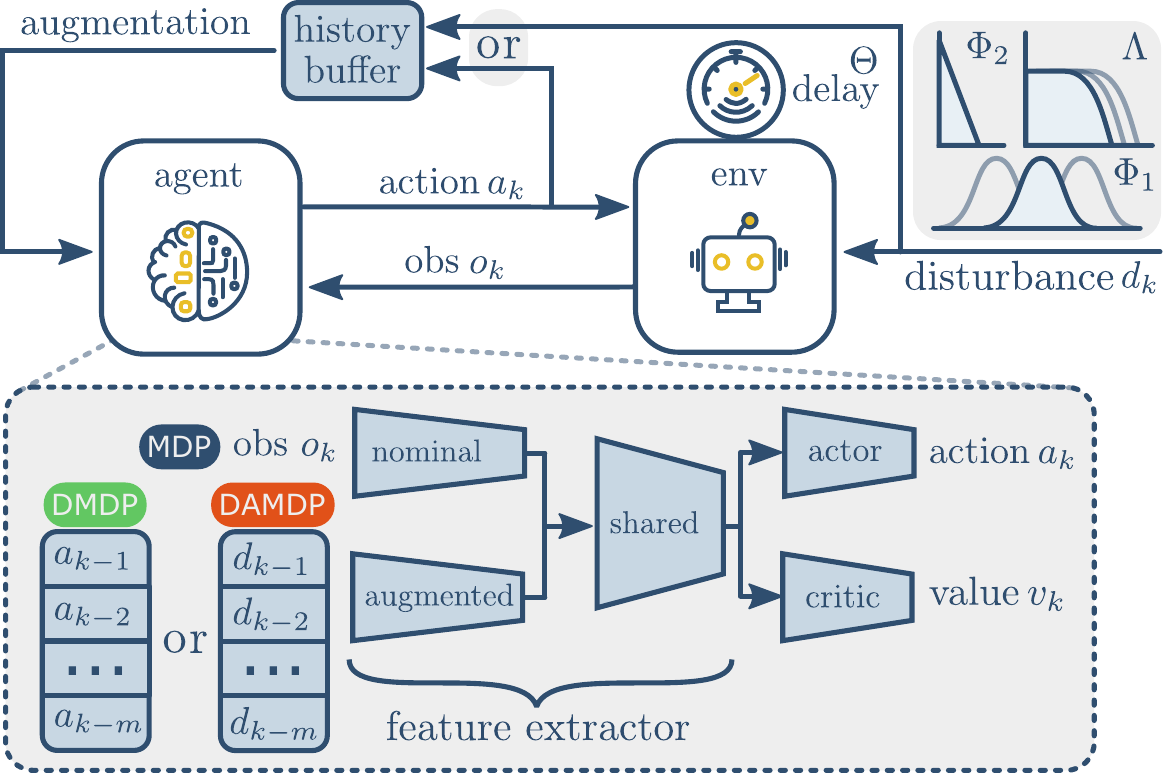}%
	\vspace{-15pt}
	\caption{Delay-resolved RL utilizes action augmentations (modeled as DMDP) to preserve the Markov property in delayed environment settings. Disturbance-aware RL extends the concept to delayed and disturbed environments by exploiting disturbance-augmented observations, where a sequence of estimated disturbances is augmented to the agent observations (modeled as DAMDP) to improve the robustness of the learned control policy.}%
	\label{fig:augmentations}%
\end{figure}%

Nevertheless, it is worthwhile to reduce the training time required to adapt the trained policies to the real robot and eventually diminish the need for such extensive, time-consuming training. Overcoming this problem requires developing methods that can enable the agent to achieve robust performance with respect to different environment dynamics and generalize its behavior across widely varying settings. By random variation of the simulation aspects during training, domain randomization \cite{Tobin.20Mar17,Tremblay.18Apr18}, and dynamics randomization \cite{Peng.2018,Tan.27Apr18,Rajeswaran.05Oct16} have been predominantly reported as successful methods in increasing the agent's ability to generalize to new environments. Incorporating dynamic uncertainties into the simulation allows the algorithm to experience a range of possible scenarios and develop a more robust control policy to handle unexpected situations in the real world \cite{Peng.2018,Christiano.11Oct16,Madebo2024}.

Randomizing the physical effects of a simulated robotic system induces structured uncertainties in the dynamics model, which in turn affect the behavior of the control system. Analogously, such deliberate changes in the environment can be viewed as interposing disturbances to the system's input, state, or dynamics. Conventionally, randomization has been used to gain control robustness throughout the training phase, whereas simplified disturbance injection is exploited in the test phase to examine the robustness of the control policy \cite{Glossop2022}. As a matter of fact, the utilization of disturbance estimation in training robust policies with randomized simulation has been less discussed in previous works \cite{Glossop2022}.

From this perspective, this work shows how an on-policy RL algorithm can use the estimation of injected disturbances to boost the policy's robustness against randomization extent. The disturbance-augmented Markov decision process (DAMDP) is proposed as a new representation to incorporate delay-resolved disturbance estimation in the agent observation space. The proposed method is validated by learning two distinct manipulation tasks: (1) target-reaching, treated as set-point stabilization under joint-space control, and (2) box-pushing, treated as goal-regulation under operational-space control. %
As illustrated in \Cref{fig:augmentations}, the method's performance is compared with two prevailing representations: (1) the vanilla representation with nominal observations framed as a normal Markov decision process (MDP), and (2) the delay-resolved representation entailing action-augmented observations conceived as a delayed Markov decision process (DMDP) \cite{Katsikopoulos2003}. The experiment results, both on the simulation and the real robot, show that the disturbance-aware agent achieves higher stabilization and robustness over the same budget of training samples.

\vspace{-0.5\baselineskip}
\section{BACKGROUND AND RELATED WORK}

\noindent A crucial factor in networked control systems and teleoperated robotics is communication latency, which introduces time delays in the feedback control loop and eventually affects the RL agent to learn or adapt its behavior \cite{Rupam2018}. The partial observability induced by receiving outdated state information leads to suboptimal actions and instability in learning \cite{SINGH1994284}. In an ordinary MDP, any delay of more than one time step disregards the Markov property since the most recent committed actions become part of the environment's current state and are not observed. The early work of Katsikopoulos \emph{et al.} \cite{Katsikopoulos2003} discussed formulating DMDPs through augmented state spaces to represent deterministic or stochastic action and observation delays. Since then, numerous works used pliant model-predictive methods to estimate undelayed observation from the queue of augmented unapplied actions \cite{Walsh2009,Schuitema2010,CHEN2021119}. On the contrary, model-free approaches relied on the action-augmented observations as forms of Markovian information state to derive delay-resolved representations \cite{Ramstedt2019,bouteiller2021,Nath2021} in a proper format for mainstream deep RL algorithms such as proximal policy optimization (PPO) \cite{Schulman.20Jul17} or soft actor-critic (SAC) \cite{Haarnoja2018}.

\begin{figure}[t]%
	\centering%
	\includegraphics[width=\columnwidth]{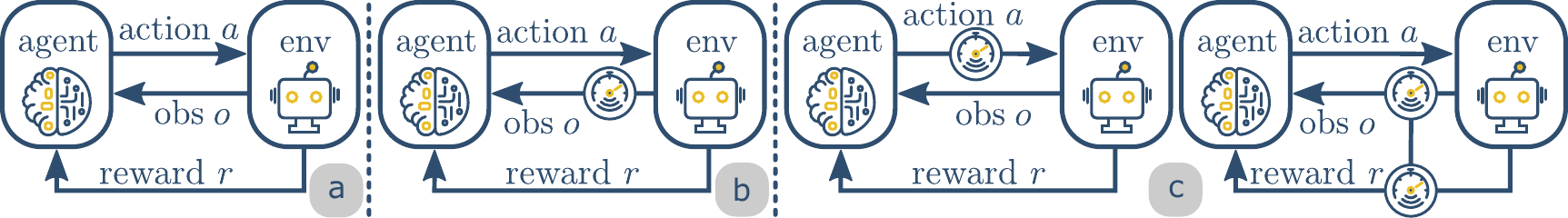}%
	\vspace{-20pt}
	\caption{Classes of delays imposed on actions, observations, and rewards: (a) no delay, (b) delayed observation with immediate reward, (c) delayed action equivalent to delayed observation with a delayed reward.}%
	\label{fig:delays}%
\end{figure}

Previous studies \cite{Katsikopoulos2003,Nath2021} have shown that action delays and observation delays are functionally equivalent from the agent's perspective. However, this argument is correct when the reward signal suffers from the same amount of delay as the observation (\Cref{fig:delays}(c); e.g., training in the real world). In randomized simulations, one can decide whether to use a delayed or immediate reward. These possible cases of delay types affecting agent-environment interactions are shown in \Cref{fig:delays}. As the first step, our study empirically shows that training with immediate rewards (\Cref{fig:delays}(b)) helps the agent converge to higher returns.

From a generic perspective of sim2real methods, stochastic network delay is just one specific type of many factors that influence the reality gap. Previous studies showed that randomization of other simulation aspects, such as parameters of actuation mechanism (e.g., motor torque profiles, joints damping, and actuation bandwidth) or physical properties of the robot and environment (e.g., mass, inertia, and friction), as well as imposing observation noise or action disturbances help in improving the robustness of the trained policies. Generally speaking, such vast dynamics randomization is supposed to help the agent learn how to adapt to different physical settings and better generalize to the real world \cite{Josifovski2022,Josifovski2024}. However, there is a big assumption in most previous works - the agent can learn better policies no matter how and to what extent it can access the environment's information state \cite{Petropoulakis2024Incentives}. Within this work, we empirically show that this assumption is not necessarily correct and strongly depends on whether or not the agent can deduce the true environment state from (partial) observations of the randomized system dynamics.

\begin{figure*}[t]%
	\centering%
	\includegraphics[width=0.98\textwidth]{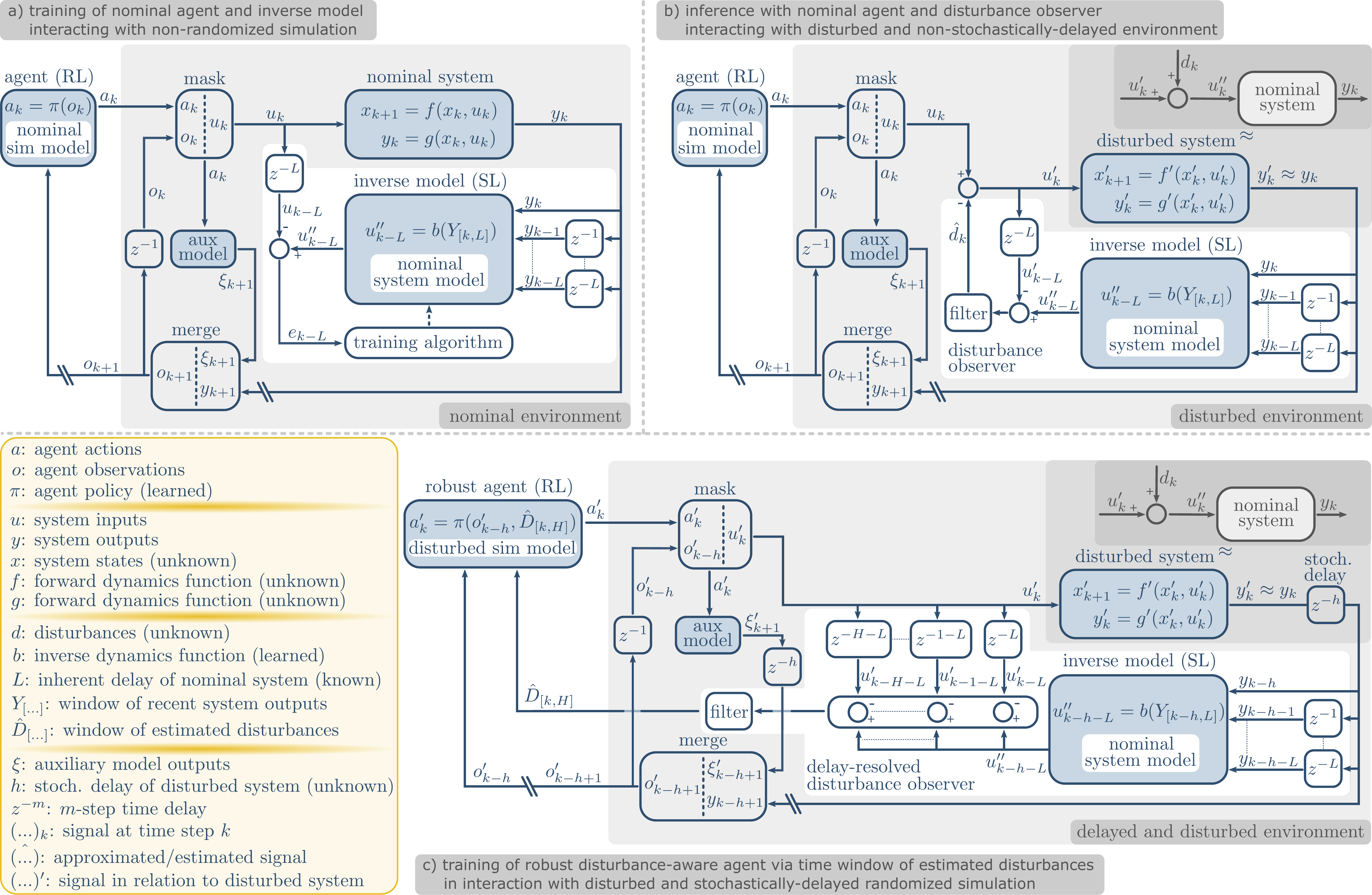}%
	\vspace{-5pt}
	\caption{(a) Illustration of how the inverse model is trained with data of a non-randomized simulation. (b) Under the assumption of known delay, DOB employs the trained inverse model to mitigate uncertainties solely via feedforwarding the latest estimated disturbance. (c) Disturbance-aware RL agent uses the delay-resolved DOB with the trained inverse model to achieve control robustness against simulated disturbances in stochastically delayed environments.}%
	\label{fig:DARL}%
	\vspace{-15pt}
\end{figure*}%

In line with the principles of $H_{\infty}$ optimal control \cite{Basar.2008}, previous studies \cite{Tan.27Apr18,LerrelPinto.2017} have addressed that uncertainties in modeling the correct dynamics of the real-world system (as a part of the reality gap) can be represented in terms of unknown disturbances added to the agent's actions. In previous works \cite{Kim.2019,Malmir2020Robust,zarei2021robotic}, the authors proposed the use of a nonlinear disturbance observer (DOB) to estimate the additive disturbances (i.e., the domain gap) between the source and target domains. According to \cite{Kim.2019,Malmir2020Robust}, incorporating the estimated disturbances to modify the agent's actions in a feedforward manner can be beneficial in ensuring that the agent is able to sustain its expected performance levels despite the presence of external disturbances or the uncertainty of the system dynamics. However, properly compensating the disturbances by a feedforward action greatly depends on accurate knowledge of the system input-output delay. In contrast to prior model-based methods, our work presents a unifying data-driven framework to integrate disturbance estimation with delay-resolved RL and train action-robust policies in randomized simulations featuring stochastic delays. We argue that disturbance-aware reinforcement learning (DiAReL) is suitable for enabling RL agents to obtain higher control robustness through adjusted policy training. Our results show that, given the agent’s delay awareness over a known horizon, incorporating disturbance awareness yields significantly higher stabilization rates than delay awareness alone.

\vspace{-0.5\baselineskip}
\section{METHODOLOGY}

\begin{algorithm}[t]
\caption{Training the Inverse Dynamics Model}
\label{alg:inv-dyn-train}
\begin{algorithmic}[1]
\Require Precollected dataset $\mathcal{D}_{\text{raw}} = \{(o_k,a_k)\}_{k=0}^K$ from nominal agent training with non-randomized simulation; Hyperparameters: learning rate $\alpha$, minibatch size $B$, epochs $E$; Nominal system config: inherent delay $L$, binary‐selection mask for inputs $M^u$ and outputs $M^y$
\Ensure Trained inverse model $b_\phi$
\State Initialize model parameters $\phi$
\State $\mathcal{D} \gets \emptyset$ \Comment{Construct system input-output pairs}
\For{$k \gets L$ to $K$}
  \If{$\mathrm{episode}(k) = \mathrm{episode}(k - L)$}
    \State $u_{k-L} \gets M^u[o_{k-L};a_{k-L}]$
    \State $Y_{[k,L]} \gets M^y[o_{k-L:k}]$ \Comment{apply $M^y$ to all the slice}
    \State Add $\bigl(Y_{[k,L]},u_{k-L}\bigr)$ to $\mathcal{D}$
  \EndIf
\EndFor
\For{epoch $e \gets 1$ to $E$}
  \State Shuffle dataset of input-output pairs $\mathcal{D}$
  \For{each minibatch $\mathcal{B}\subset\mathcal{D}$ of size $B$}
    \State Extract $\bigl\{\bigl(Y_{[k,L]},u_{k-L}\bigr)\bigr\}$ from $\mathcal{B}$
    \State Compute predictions: $u''_{k-L} = b_\phi(Y_{[k,L]})$
    \State Compute loss: $\mathcal{L} \gets \tfrac{1}{B}\sum_{\mathcal{B}}\|u''_{k-L} - u_{k-L}\|_2^2$
    \State Update $\phi \gets \phi - \alpha\nabla_\phi \mathcal{L}$
  \EndFor
\EndFor
\State \Return $b_\phi$
\end{algorithmic}
\end{algorithm}

\begin{algorithm}[t]
\caption{Disturbance-Aware On-Policy RL}
\label{alg:diarel}
\begin{algorithmic}[1]
\Require Disturbed and stochastically-delayed environment; Hyperparameters: rollout length $T$, minibatch size $B$, epochs $E$; DOB config: Delay horizon $H$, time step $T_s$, cutoff freq. $f_c$, inherent delay $L$, trained inverse model $b_\phi$, binary‐selection mask for inputs $M^u$ and outputs $M^y$
\Ensure Learned policy $\pi_\theta(a' \mid \prescript{+d}{}{o})$
\State Initialize policy parameters $\theta$
\State Compute filter bandwidth coefficient $\beta \gets \exp\bigl(-2\pi f_c T_s\bigr)$
\For{iteration $\gets 1$ to max\_iters}
  \State Initialize all $\hat{d}_{j} \gets 0$ for $j = -H,\dots,0$
  \For{$k \gets 1$ to $T$} \Comment{Collect rollout of $T$ steps}
    \State Observe $o'_{k-h}$ with unknown delay $h \in [0,H]$
    \If{$k > L$}
      \State $Y_{[k-h,L]} \gets M^y[o'_{k-h-L : k-h}]$  
      \State $u''_{k-h-L} \gets b_\phi(Y_{[k-h,L]})$
    \EndIf
    \For{$i \gets 0$ to $H$}
      \State $\hat{d}_{k-i} \gets \mathbf{0}$
      \If{$\mathrm{episode}(k) = \mathrm{episode}(k - i - L)$}
        \State $u'_{k-i-L} \gets M^u[o'_{k-h-i-L};a'_{k-i-L}]$        
        \State $\delta \gets u''_{k-h-L} - u'_{k-i-L}$
        \State $\hat{d}_{k-i} \gets \delta + \beta \bigl(\hat{d}_{k-i-1} - \delta\bigr)$
      \EndIf
    \EndFor
    \State Form $\prescript{+d}{}{o}_{k} \gets \bigl(o'_{k-h},[\hat{d}_{k},\dots,\hat{d}_{k-H}]^T\bigr)$
    \State Sample action $a'_k \sim \pi_\theta(\cdot \mid \prescript{+d}{}{o}_{k})$
    \State Execute $a'_k$, receive reward $r_k$ and next observation
  \EndFor
  \State Compute value targets $\hat{V}_k$ from $\{(\prescript{+d}{}{o}_{k},a'_k,r_k)\}_{k=1}^T$
  \State Compute advantages $\hat{A}_k$ from $\{(\prescript{+d}{}{o}_{k},a'_k,r_k)\}_{k=1}^T$
  \State Form rollout buffer $\mathcal{D} \gets \{(\prescript{+d}{}{o}_{k},a'_k,\hat{A}_k,\hat{V}_k)\}_{k=1}^T$
  \For{epoch $e \gets 1$ to $E$}
    \State Shuffle $\mathcal{D}$
    \For{each minibatch $\mathcal{B} \subset \mathcal{D}$ of size $B$}
      \State Update $\theta$ via actor‐critic loss on $\mathcal{B}$
    \EndFor
  \EndFor
\EndFor
\State \Return $\pi_\theta$
\end{algorithmic}
\end{algorithm}

\noindent This work introduces a control framework for designing and training data-driven nonlinear DOB, seamlessly integrated with delay-resolved RL settings. Having a nominal dynamical system (as part of the simulated environment) with inputs $u_k$ and outputs $y_k$, a DOB uses a model-based or pre-trained inverse model to compute the input $u_{k-L}$ imposed on the nominal system using a fixed time window of recent outputs $Y_{[k,L]}=[y_{k},y_{k-1},\ldots,y_{k-L}]^T$ (\Cref{fig:DARL}(a)). $L$ is a predetermined constant integer representing the inherent delay of the non-randomized simulation. To train the inverse model, the nominal agent interacts with the non-randomized simulation and follows the principle of a standard MDP to learn an optimal policy. The labeled data for training the inverse model in a supervised learning (SL) fashion is sampled from the pre-collected explorative sequences that the nominal agent experiences during its training. The inverse model learns the inverse dynamics of the nominal system, a division of the non-randomized environment that will be prone to disturbances at the inference time. This division is achieved by selectively choosing specific system inputs $u_k$ from the stacked vector of agent observations and actions $[o_k;a_k]$. The selection process is controlled by the mask $M^u$, which is a design parameter tailored to the specific requirements of the task. The auxiliary model (aux model at \Cref{fig:DARL}(a)) captures the complementary portion of the environment dynamics, which is characterized by low uncertainty and thus is not relevant to the DOB. The steps of inverse model training are shown in \Cref{alg:inv-dyn-train}.

In disturbed environments without stochastic delay, DOB uses the trained inverse model to estimate the uncertainties raised by randomization and applies a feedforward action (\Cref{fig:DARL}(b)). Here, the constant system input-output delay is known beforehand. When the agent-environment interaction suffers from stochastic delays, an accurate estimation of the delay is necessary for proper disturbance compensation. Inspired by delay-resolved RL algorithms \cite{Ramstedt2019,bouteiller2021,Nath2021}, \Cref{fig:DARL}(c) shows our schematic design of the training loop for the disturbance-aware agent in the presence of unknown stochastic delays. A sequence of estimated disturbances is augmented to the agent observations to improve the robustness of the learned control policy and thus effectively transfer it from simulation to the real world. DAMDP is proposed as a concept to train robust agents with delay-resolved disturbance awareness.

\begin{definition}[DAMDP]
An m-stage disturbance-augmented Markov decision process extends a conventional Markov decision process $\text{MDP}(S,A,P,r,\gamma)$ with state space $S$, action space $A$, state-transition probability matrix $P$, reward function $r$ and discount factor $\gamma$ to a $\text{DAMDP}(\prescript{+d}{}{S},A,\prescript{+d}{}{P},\prescript{+d}{}{r},\gamma)$, where $\prescript{+d}{}{S}=S\times D^{m}$ is the disturbance-augmented state space, $D$ is the vector space of estimated disturbances, $m$ is the length of the time window, and $\prescript{+d}{}{P}$ and $\prescript{+d}{}{r}$ are correspondingly the state-transition probability matrix and the reward function on the augmented state space $\prescript{+d}{}{S}$.
\end{definition}

As depicted in \Cref{fig:DARL}(c), the augmented state space $\prescript{+d}{}{S}$ includes a finite window of disturbances estimated by the pre-trained nonlinear DOB. At each time step $k$, the agent receives a disturbance-augmented observation $\prescript{+d}{}{o}_{k}=(o^\prime_{k-h},\hat{D}_{[k,H]})$, where $o^\prime_{k-h}$ is the latest observation delayed by $h \in [0,H]$ time steps, and $\hat{D}_{[k,H]}=[\hat{d}_{k},\hat{d}_{k-1},\ldots,\hat{d}_{k-H}]^T$ are the estimated disturbances across the delay horizon $H=m-1$. It is assumed that the augmented agent observation $\prescript{+d}{}{o}$ can correctly represent the DAMDP's state $\prescript{+d}{}{s}$ since full observability has been gained through the delay-resolved setting. Based on this observation, the agent decides on an action $a'_{k}$ and changes the controlled input $u'_{k}$ to the uncertain system. \Cref{alg:diarel} indicates how the DiAReL framework uses the trained inverse model to integrate a disturbance observer into an on‐policy RL loop under the DAMDP formulation.

\section{EXPERIMENT SETUP}

\begin{figure}[b]%
	\centering%
	\includegraphics[width=\columnwidth]{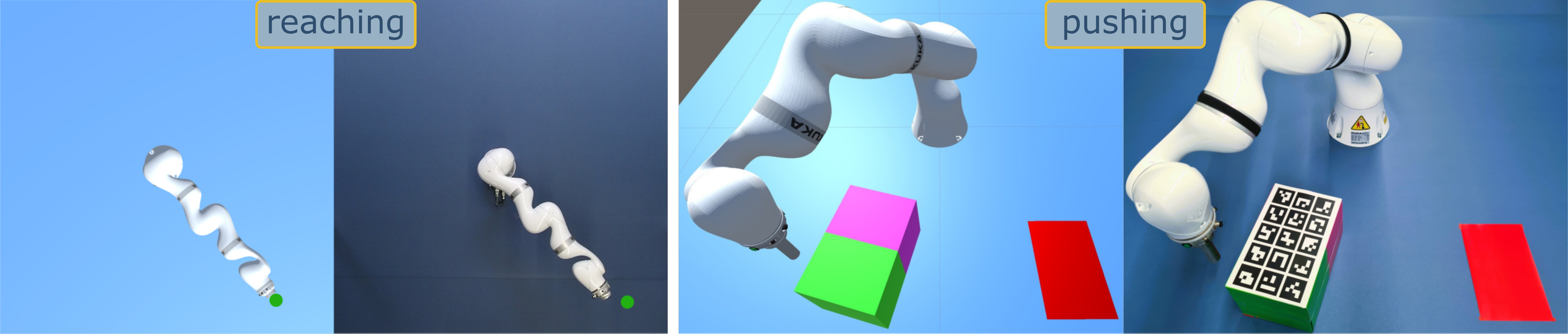}%
	\vspace{-20pt}
	\caption{(Left: Simulation, Right: Real) images of the environments with the Kuka LBR iiwa 14 for (a) the reaching task to a green target with the robot's two first joints controlled, and (b) the box pushing task to a red placement target with task space planar control of the pushing pin. In pushing, the box's center of mass lies at the green section's midpoint, showing the purple section is lighter. To estimate the box pose, an ArUco board \cite{garrido2014automatic} has been used.}
	\label{fig:sim_and_real}%
	\vspace{-5pt}
\end{figure}

\noindent The simulation environment\footnote[1]{The simulator used for the experiments is available at: https://github.com/tum-i6/VTPRL/tree/pushing-task-updates}, developed with Unity and Dynamic Animation and Robotics Toolkit (DART), provides an interface similar to OpenAI Gym and supports parallel robot simulations to accelerate training. The environment comprises a floor-mounted KUKA LBR iiwa 14 robot manipulator. The robot's parameters and meshes used in the simulation were obtained from the URDF data provided by ROS-Industrial, whereas the real robot was controlled using the IIWA stack through ROS. \Cref{fig:sim_and_real} shows top views of the simulated and the real environments in the two tasks: (a) target reaching and (b) box pushing.

The performance of disturbance-aware agents in learning the above tasks in delayed environments is compared with agents using vanilla representation with no augmentation $\text{MDP}(S,A,P,r,\gamma)$ and delay-resolved agents using action-augmented observations $\text{DMDP}(\prescript{+a}{}{S},A,\prescript{+a}{}{P},\prescript{+a}{}{r},\gamma)$. The delay-resolved agents have the augmented state space of $\prescript{+a}{}{S}=S\times A^{m}$.

We used PPO (StableBaselines PyTorch implementation \cite{Raffin2021StableBaselines3}) with a modified feature extractor for all three representations. As shown in \Cref{fig:augmentations}, the features are extracted using separate layers for the nominal and augmented parts of the observation vector. The network architecture of the augmented models represents mid-fusion to a shared layer, followed by separate heads for the actor and critic. The hyperparameters of PPO with generalized advantage estimation (GAE) \cite{schulman2018highdimensional} (\Cref{table:hyperparameters_agent}) were first seeded from Henderson \textit{et al.} \cite{Henderson2018} and then tuned via Optuna on the nominal reaching and pushing tasks, with the final values fixed across all robust variants. The inverse model is structured as a multilayer perceptron (MLP) and trained with mean square error (MSE) as the loss function. DOB hyperparameters (\Cref{table:hyperparameters_dob}) were similarly initialized via Optuna‐derived defaults and refined locally on validation error.

Experiment 1 (\textit{reaching}): The experiment involves a reach-and-balance task similar to \cite{Josifovski2022}. The agent learns the differential inverse kinematics (IK) of the two controlled links to learn to reach the target location via the robot's end effector and maintain stability at the specified target until the end of the episode, which is crucial for real-world safety. The agent actions are
\begin{equation}
a = [\dot {q}_1^{\prime},\dot{q}_2^{\prime}]^T \in \mathbb{R}^2 \text{,}
\end{equation}
\noindent where $\dot {q_i}^{\prime} \in [-1,+1]; \forall i$ are the commanded continuous joint velocities in normalized quantities and have correspondence with the system input $u$. The system output $y$ matches the observation coming from the nominal environment and is
\begin{equation}
o = [\delta_{x}, \delta_{y}, \delta_{z},{q}_1,{q}_2]^T \in \mathbb{R}^5 \text{,}
\end{equation}
\noindent where $\delta_{j} \propto t_j-e_j$ are the normalized errors between the Cartesian coordinates of the target location $t$ and the end effector location $e$, and ${q_i}$ are the normalized observed joint positions. The reward function is defined as
\begin{equation}
{r_k} = 
    \begin{Vmatrix}
    \delta_{x_{k-1}}\\
    \delta_{y_{k-1}}\\
    \delta_{z_{k-1}}
    \end{Vmatrix}_2 - 
    \begin{Vmatrix}
    \delta_{x_{k}}\\
    \delta_{y_{k}}\\
    \delta_{z_{k}}
    \end{Vmatrix}_2
    \in \mathbb{R} \text{,}
\end{equation}
\noindent where $k \in [1,N]$ is the episode time step ($r_0=0$), and $N=10^3$ is the episode duration.

Experiment 2 (\textit{pushing}): The experiment involves pushing a bulky, rigid, box-shaped object with dimensions $(0.3,0.2,0.2)$ meters toward a desired configuration on a horizontal plane, and it can be framed as a non-prehensile dynamic manipulation task \cite{push_ruggiero}. In nominal settings, the object has a uniform mass distribution, and its geometric center represents the center of mass. Friction and contact forces play a pivotal role in changing the motion of the pushed object. Simulating these factors accurately presents a challenge, amplifying the uncertainty within the modeled system \cite{push_stueber}. The agent actions are
\begin{equation}
a = [\dot {e}_x^{\prime},\dot{e}_y^{\prime}]^T \in \mathbb{R}^2 \text{,}
\end{equation}
\noindent where $\dot {e_i}^{\prime} \in [-1,+1]; \forall i$ are the commanded continuous end effector velocities within the Cartesian task space in normalized quantities. The observation coming from the nominal environment is the vector
\begin{equation}
o = [\delta_{p_\theta}, \delta_{p_x}, \delta_{p_y}, \delta_{r_x}, \delta_{r_y}]^T \in \mathbb{R}^5 \text{,}
\end{equation}
\noindent where $\delta_{p_j} \propto t_j-o_j$ are the normalized errors between the yaw angle $\theta$ or Cartesian coordinates $x,y$ of the target location $t$ and the object location $o$, and $\delta_{r_j} \propto o_j-e_j$ are the normalized errors between the Cartesian coordinates of the object location $o$ and the end effector location $e$. The reward function is defined as
\begin{equation}
{r_k} = 
    \begin{cases}
    \begin{Vmatrix}
    \delta_{r_{x_{k-1}}}\\
    \delta_{r_{y_{k-1}}}
    \end{Vmatrix}_2 - 
    \begin{Vmatrix}
    \delta_{r_{x_{k}}}\\
    \delta_{r_{y_{k}}}
    \end{Vmatrix}_2 , \text{ if object is not moved} \\
    w_k (|\delta_{p_{\theta_{k-1}}}| - |\delta_{p_{\theta_{k}}}|) + 
    \begin{Vmatrix}
    \delta_{p_{x_{k-1}}}\\
    \delta_{p_{y_{k-1}}}
    \end{Vmatrix}_2 - 
    \begin{Vmatrix}
    \delta_{p_{x_{k}}}\\
    \delta_{p_{y_{k}}}
    \end{Vmatrix}_2 , \text{ ow}
    \end{cases}
\end{equation}
\noindent where $w_k$ is a weighting factor, $k \in [1,N-1]$ is the episode time step and $N=300$ is the episode duration. The reward at the first step $r_0$ is zero, and at the final step $r_N$ is the above formulation plus the object placement accuracy, which is the weighted distance in $\theta/x,y$ from the object to the target. The system under uncertainty in \textit{pushing} is considered the object, not the robot. Hence, the system inputs and outputs differ from the environment actions and observations and are $u = [\delta_{r_x}, \delta_{r_y}]^T$ and $y = [\delta_{p_\theta}, \delta_{p_x}, \delta_{p_y}]^T$ respectively.

\begin{table}[t]
    \begin{center}
    \caption{Nominal/Robust Agent Hyperparameters (PPO).}
    \vspace{-5pt}
    \label{table:hyperparameters_agent}
    \begin{tabular}{ll}
    \hline\noalign{\smallskip}
     Feature extractor  &  $ \text{MLP} \begin{pmatrix}
                                            \text{nominal: }-\\
                                            \text{augmented: }64\\
                                            \text{shared: }-
                                            \end{pmatrix} $ [Tanh] \vspace{3pt} \\
     Policy network structure  &  MLP $(64,64)$ [Tanh] \\
     Critic network structure  &  MLP $(64,64)$ [Tanh] \\
     Number of actors  & $64$ \\
     Rollout length ($T$)  & reaching: $256$ / pushing: $512$ \\
     Minibatch size ($B$)  &  $512$ \\
     Number of epochs ($E$)  &  $10$ \\
     Learning rate ($\alpha$)  & $2.5\times10^{-4}$ \\
     Discount factor ($\gamma$)   &   $0.99$ \\
     GAE parameter ($\lambda$)  &  $0.95$ \\
     Clipping parameter ($\epsilon$)  &  $0.1$ \\
     Value function coefficient ($c_{1}$)  &  $0.5$ \\
     Entropy coefficient ($c_{2}$) & $0.0$ \\
    \hline\noalign{\smallskip}
    \end{tabular}
    \end{center}
    \vspace{-15pt}
\end{table}

\begin{table}[t]
    \begin{center}
    \caption{Disturbance Observer Hyperparameters (DOB).}
    \vspace{-5pt}
    \label{table:hyperparameters_dob}
    \begin{tabular}{ll}
    \hline\noalign{\smallskip}
     Inverse model network structure  &  MLP $(64,64)$ [Tanh] \\
     Discrete time step ($T_s$)  &  reaching: $0.02\,$s / pushing: $0.1\,$s \\
     Cutoff frequency ($f_c$)  &  $1.0\,$Hz \\
     Nominal system delay ($L$)  & $1$ \\
     Input mask ($M^u$)  & reaching: $[0_{2\times 5}\;I_{2\times 2}]$; \\
       & pushing: $[0_{2\times 3}\;I_{2\times 2}\;0_{2\times 2}]$ \\
     Output mask ($M^y$)  & reaching: $I_{5\times 5}$; \\
       & pushing: $[I_{3\times 3}\;0_{3\times 2}]$ \\
     Delay horizon ($H$)  & reaching: $14$ / pushing: $2$ \\
     Minibatch size ($B$)  &  $128$ \\
     Number of epochs ($E$)  &  $5$ \\
     Learning rate ($\alpha$)  & $1.0\times10^{-3}$ \\
     Training-validation split ratio  &  $T:90\% - V:10\%$ \\
    \hline\noalign{\smallskip}
    \end{tabular}
    \end{center}
    \vspace{-10pt}
\end{table}

In case of action or disturbance augmentations, the agent receives an extended observation vector that appends $[{a}_{k},{a}_{k-1},\ldots,{a}_{k-H}]^T$ or $[\hat{d}_{k},\hat{d}_{k-1},\ldots,\hat{d}_{k-H}]^T$ to the nominal observation $o$. In our experiments, we consider $f=50$ Hz or $f=10$ Hz, respectively, for the \textit{reaching} or \textit{pushing}\footnote[2]{A lower sampling frequency for \textit{pushing} is used so that each action yields a sufficiently large end effector displacement, while keeping commanded speeds low to prevent uncontrolled sliding (e.g., "kicking" the box).} as the sampling frequency for the control loop in the simulation and real setup. The real-world communication latency is assumed to not exceed $\xi=0.3$ s, because of which we set the delay horizon $H=m-1=\xi \times f - 1$ as $14$ time steps for the \textit{reaching} and $2$ time steps for the \textit{pushing}.

Each model is trained for $K=10^7$ time steps in total and evaluated at every $10^5$ time step on a few customized metrics for a fixed set of evaluation episodes. For \textit{reaching}, $50$ reachable targets are considered for evaluating the performance of the models based on three defined metrics. The first metric represents the average distance to reach the target and is defined as negated position distance (NPD) in meters,
\begin{equation}
\text{NPD}\,[\text{m}] = -\frac{1}{N} \sum_{k=0}^{N}
    \begin{Vmatrix}
    \delta_{x_{k}}\\
    \delta_{y_{k}}\\
    \delta_{z_{k}}
    \end{Vmatrix}_2 \text{.}
\end{equation}
\noindent The non-normalized version of NPD has been used in previous works \cite{Josifovski2022} as a reward for reaching tasks. In contrast, rewarding the agent by the chosen distance displacements $r_k$ (as in \cite{Katyal2017,Petropoulakis2024Incentives}) showed a faster convergence rate in training. The next metric evaluates the policy on the time it takes to reach the random target. It resembles the rise time of a controlled system and is defined as reversed rise time (RRT) in seconds,
\begin{equation}
\text{RRT}\,[\text{s}] = \frac{1}{f} (N - k_r) \text{,}
\end{equation}
\noindent where $k_r \in [0,N]$ is the first time step that the distance to the target becomes less than $e_r=0.05$ m. The third metric evaluates the policy in regulating a stable, non-oscillating motion around the target. It indicates the integral of the absolute error in the vicinity of the target after the rise time step and is defined as stabilization strength (SS) in percentage,
\begin{equation}
\text{SS}\,[\text{\%}] =
    \frac{1}{N-k_r} \sum_{k=k_r}^{N} \max \left \{0, 1 - \frac{1}{e_r}
    \begin{Vmatrix}
    \delta_{x_{k}}\\
    \delta_{y_{k}}\\
    \delta_{z_{k}}
    \end{Vmatrix}_2 \right\} \text{.}
\end{equation}

In the case of \textit{pushing}, the considered metric is a slightly modified version of NPD, which considers the average distance in pushing the object toward the target in $x$ and $y$ directions. The efficacy of the three policy architectures (vanilla, action-augmented, and disturbance-augmented) in finding a robust and optimal policy for the two tasks has been evaluated by performing several training runs that feature randomized simulations with different complexities. In the case of \textit{reaching}, a model-based solution by a proportional controller with inverse kinematics (P-IK), which preserves the Markov property\footnote[3]{P-IK is used as a baseline because PI-IK with integral action would store the cumulative error over all past steps (making it non-Markovian), whereas our DMDP/DAMDP remain Markovian by using only a fixed history window.}, has been considered to represent a baseline performance for all the metrics. Three types of randomization have been considered and selectively applied during the training and evaluation.

1) Sampled stochastic delay with set $\Theta$: The real-world latency is assumed to be quasi-static and unknown within a certain range $[0,\xi]$ s. Thus, the discrete simulated delay is randomly sampled from the uniform distribution $U(0,m)$ at the beginning of each episode and kept constant during the episode. This delay can be enforced either on actions (\Cref{fig:delays}(c)) or observations (\Cref{fig:delays}(b)).

2) Actuator bandwidth limitation with set $\Lambda$: The simulated robot motors represent unrealistically high torques in rotating the joints and lack a correct identification from the real robot. The real robot has limited actuation and control bandwidth. This fact is handled in the simulation by applying a first-order low-pass filter with a random cut-off frequency $f'_c \thicksim U(0.2,2.0)$ on the actions.

3) External/Internal disturbance with set $\Phi$: This randomization is task-dependent and either imposed on the actions or physical parameters of the components existing in the environment. In \textit{reaching}, the disturbances are additive to the actions and considered either biased stochastic disturbances $\Phi_1$ or step disturbances $\Phi_2$. The biased stochastic disturbances are constantly applied as a Gaussian $N(\mu,\sigma^2)$ with standard deviation $\sigma=0.2$ and a varying mean that is sampled from the uniform distribution $U(-0.7,0.7)$ at the beginning of each episode. The step disturbances have a constant norm of $0.5$ for each action with sign randomization, i.e., $d_k=(\pm 0.5, \pm 0.5)$. The step disturbances \cite{Glossop2022} are injected either from the episodes' beginning $k=0$ (representative of systematic errors in actuation) or from the middle of each episode $k=500$ (representative of external disturbances imposed on the real robot) and kept constant during the episode. In randomized \textit{pushing}, the box has an unknown mass distribution across its length, acting as internal disturbances $\Phi_3$. Each episode simulates a randomized object where the object's center of mass is initialized to $\{-0.075,0.0,+0.075\}$ meters apart from the object's midpoint.

\vspace{-0.5\baselineskip}
\section{EXPERIMENT RESULTS}

\begin{figure}[b]%
	\vspace{-5pt}
	\centering%
	\includegraphics[width=0.25\columnwidth]{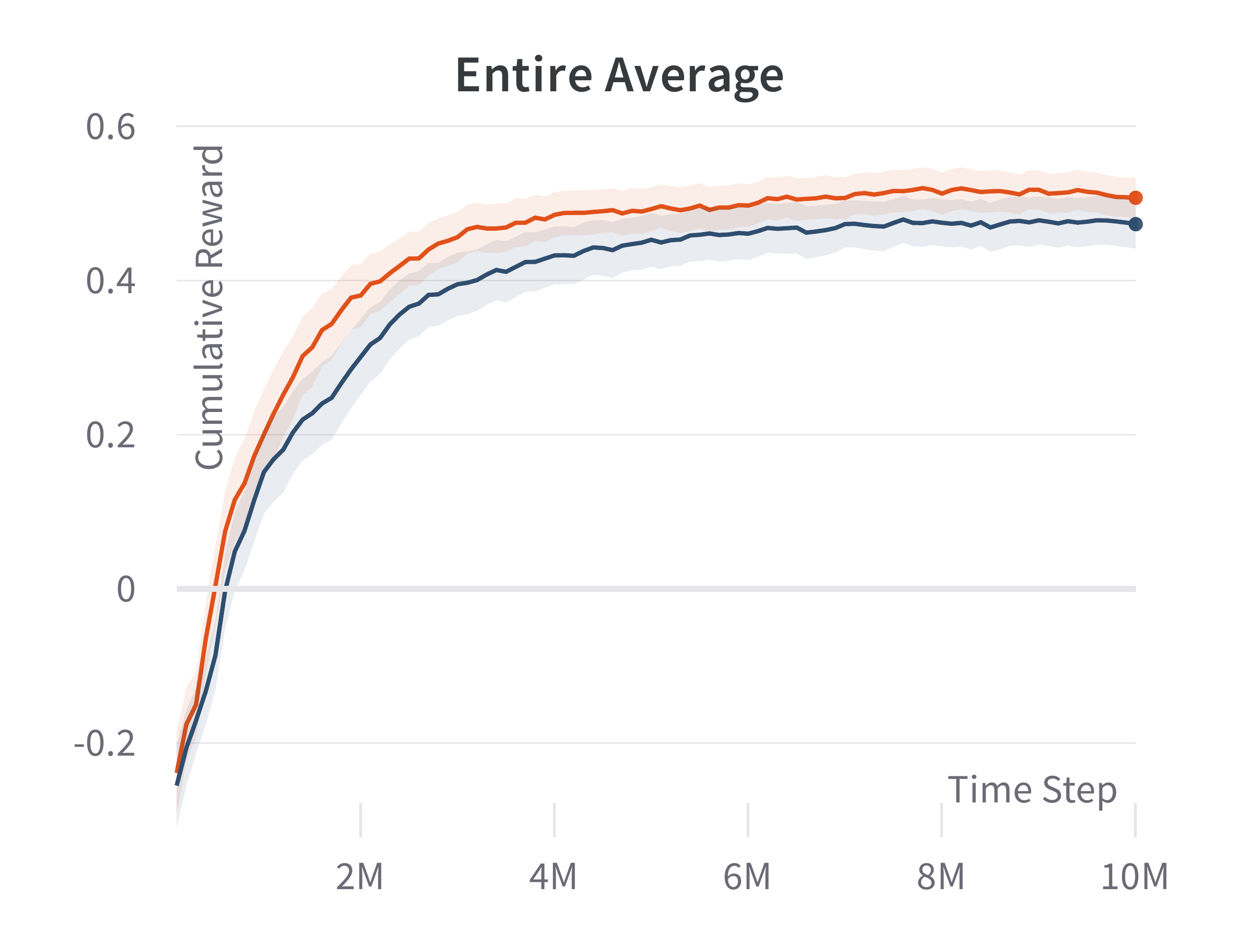}%
	\includegraphics[width=0.25\columnwidth]{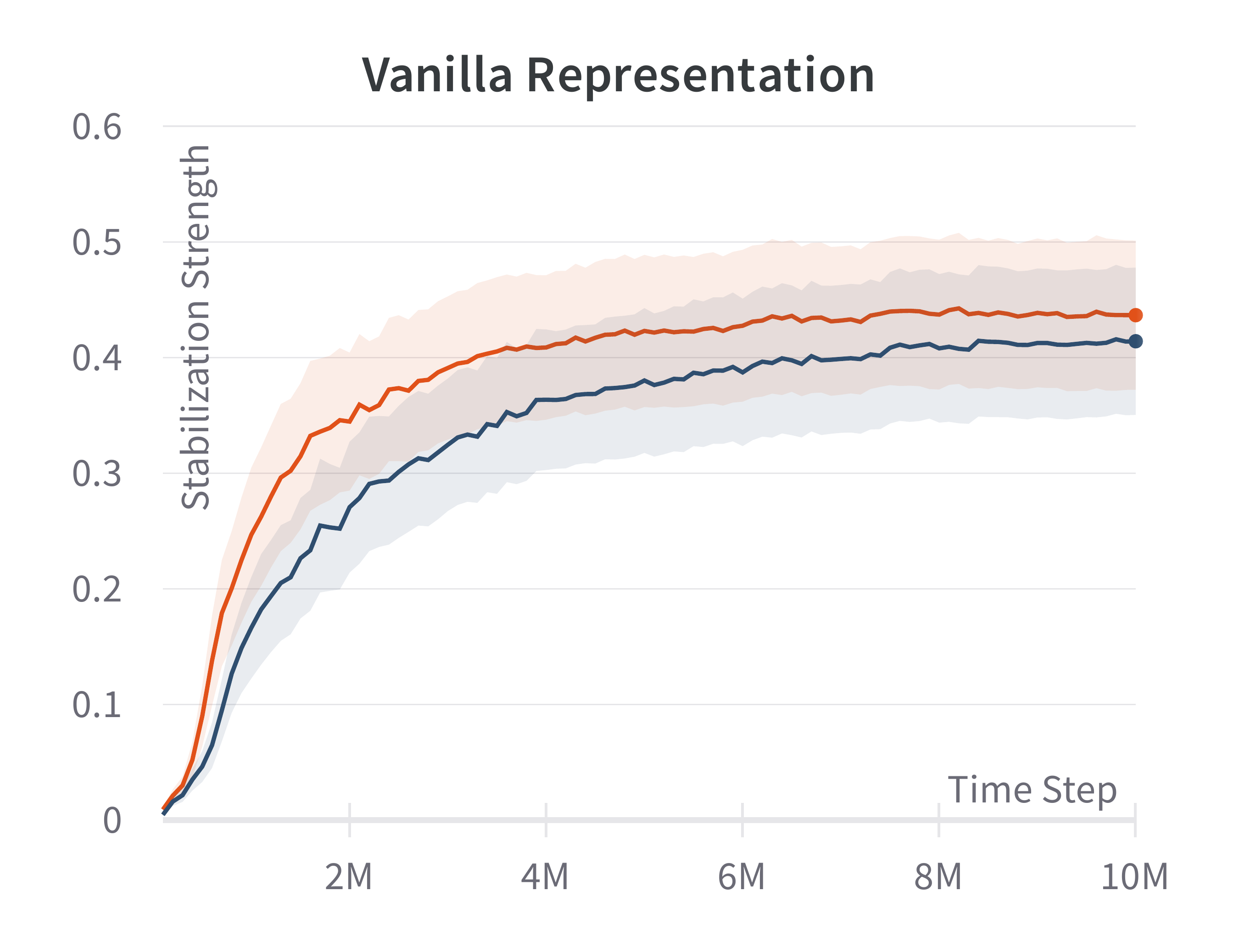}%
	\includegraphics[width=0.25\columnwidth]{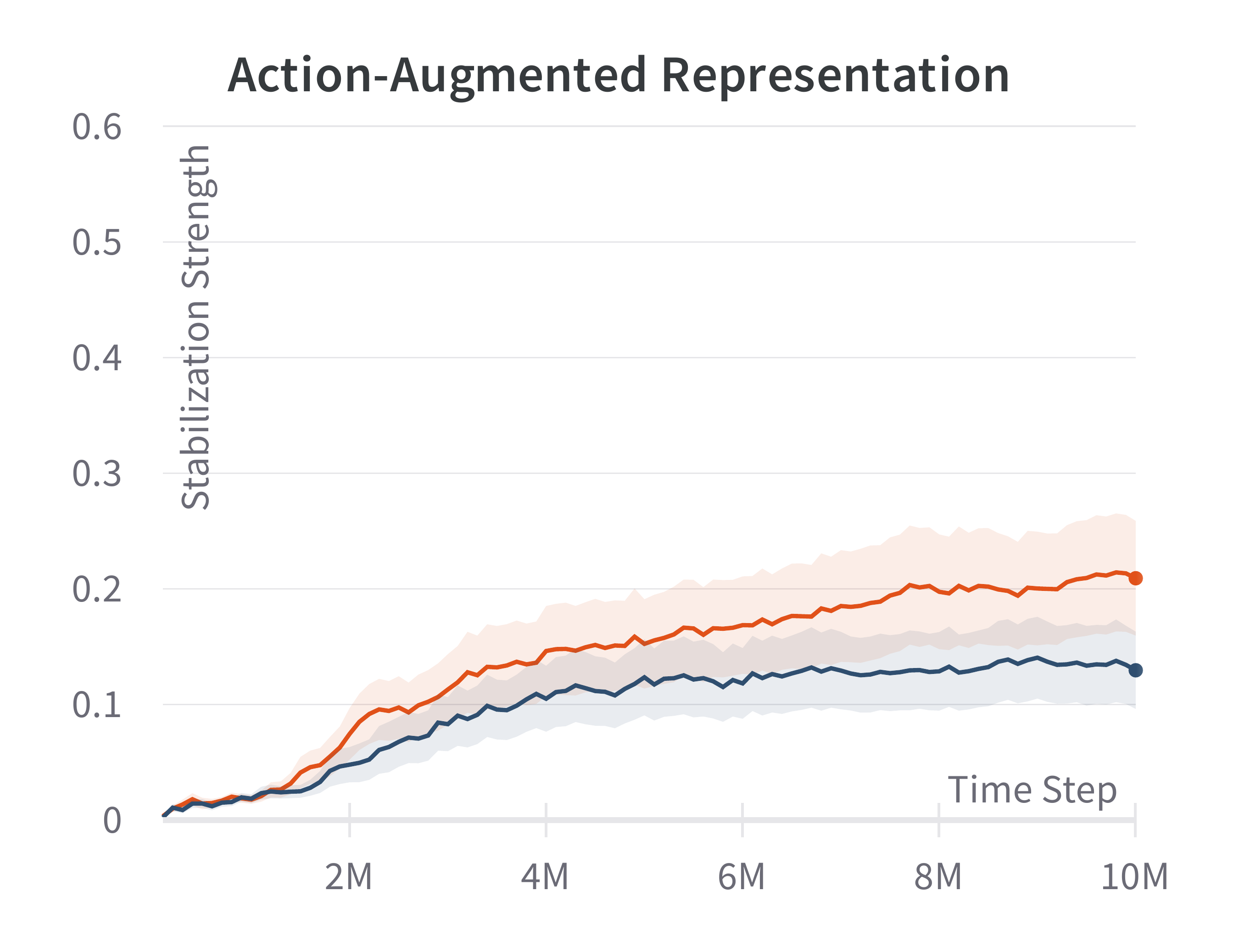}%
	\includegraphics[width=0.25\columnwidth]{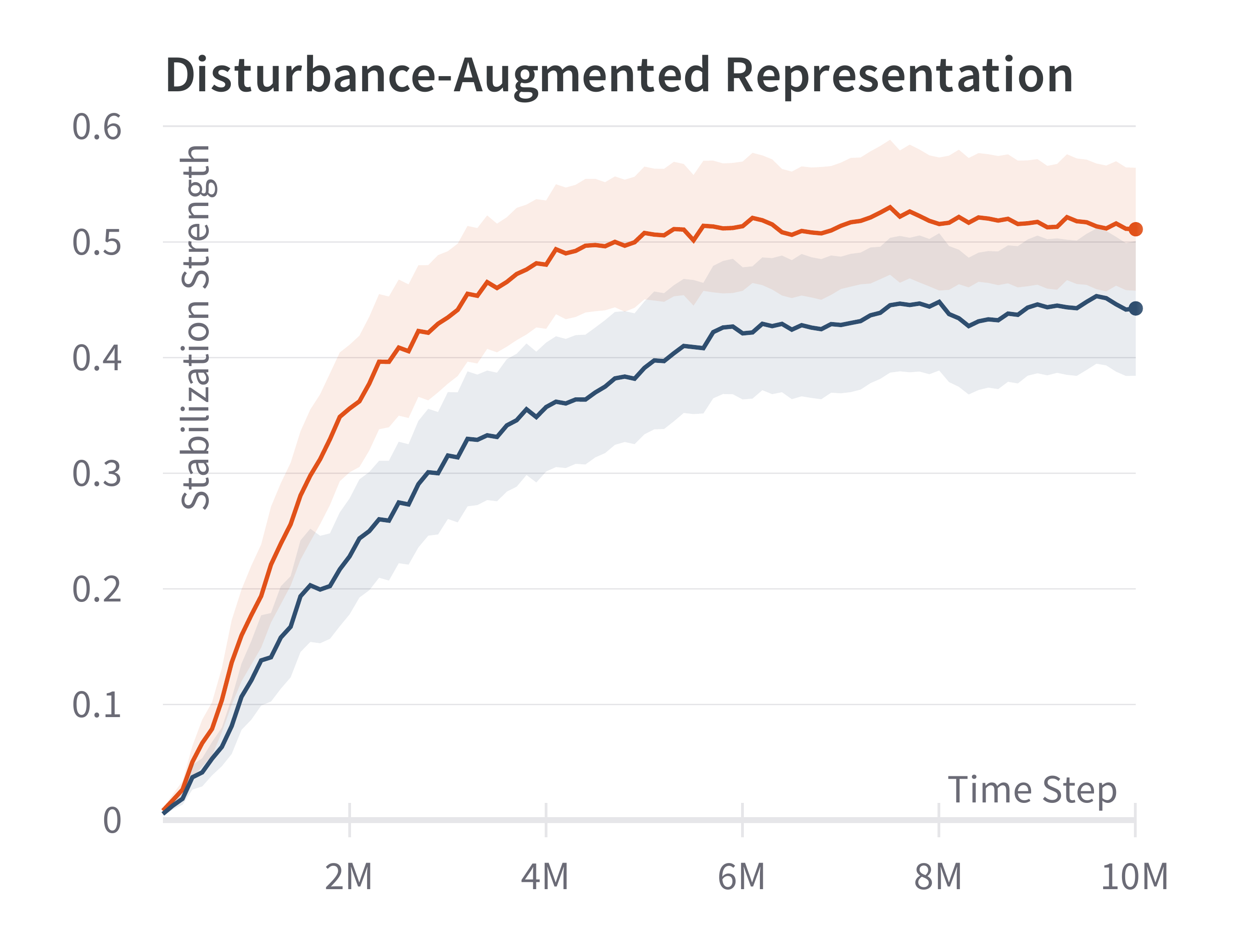}\\%
	\vspace{-10pt}
    \makebox[\columnwidth][r]{%
	   \includegraphics[width=0.25\columnwidth]{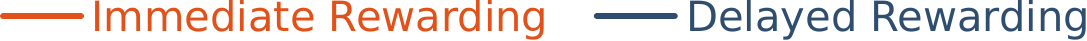}%
    }
	\vspace{-20pt}
	\caption{Impact of immediate rewarding vs. delayed rewarding on the \textit{reaching} experiment for the three distinct representations and the entire average over them. Results are averaged across five random seeds and three randomization sets $\Omega_1 \cup \Omega_2 \cup \Omega_3$. Shaded areas denote the standard error of the mean.}%
	\label{fig:delayed_reward}%
\end{figure}%

\noindent We first investigate the effect of stochastic delays on the reward signal using the \textit{reaching} experiment. Robust models were trained across five random seeds under three variations of the randomization set $\Omega$: 
\begin{enumerate}
    \item sole delay randomization $\Omega_1 = \Theta$, 
    \item delay and bandwidth randomization $\Omega_2 = \Theta \times \Lambda$, 
    \item all three types of randomization $\Omega_3 = \Theta \times \Lambda \times \Phi_1$. 
\end{enumerate}
\Cref{fig:delayed_reward} compares models trained with immediate rewarding (\Cref{fig:delays}(b)) versus delayed rewarding (\Cref{fig:delays}(c)). Simulation results show that by immediate rewarding, the same policy networks consistently yield higher cumulative rewards and better stabilization strength across all three representation types. Furthermore, models with disturbance-augmented representation outperform both action-augmented and vanilla representations in stabilization strength, suggesting their enhanced robustness.

\begin{figure}[t]%
	\centering%
	\includegraphics[width=0.33\columnwidth]{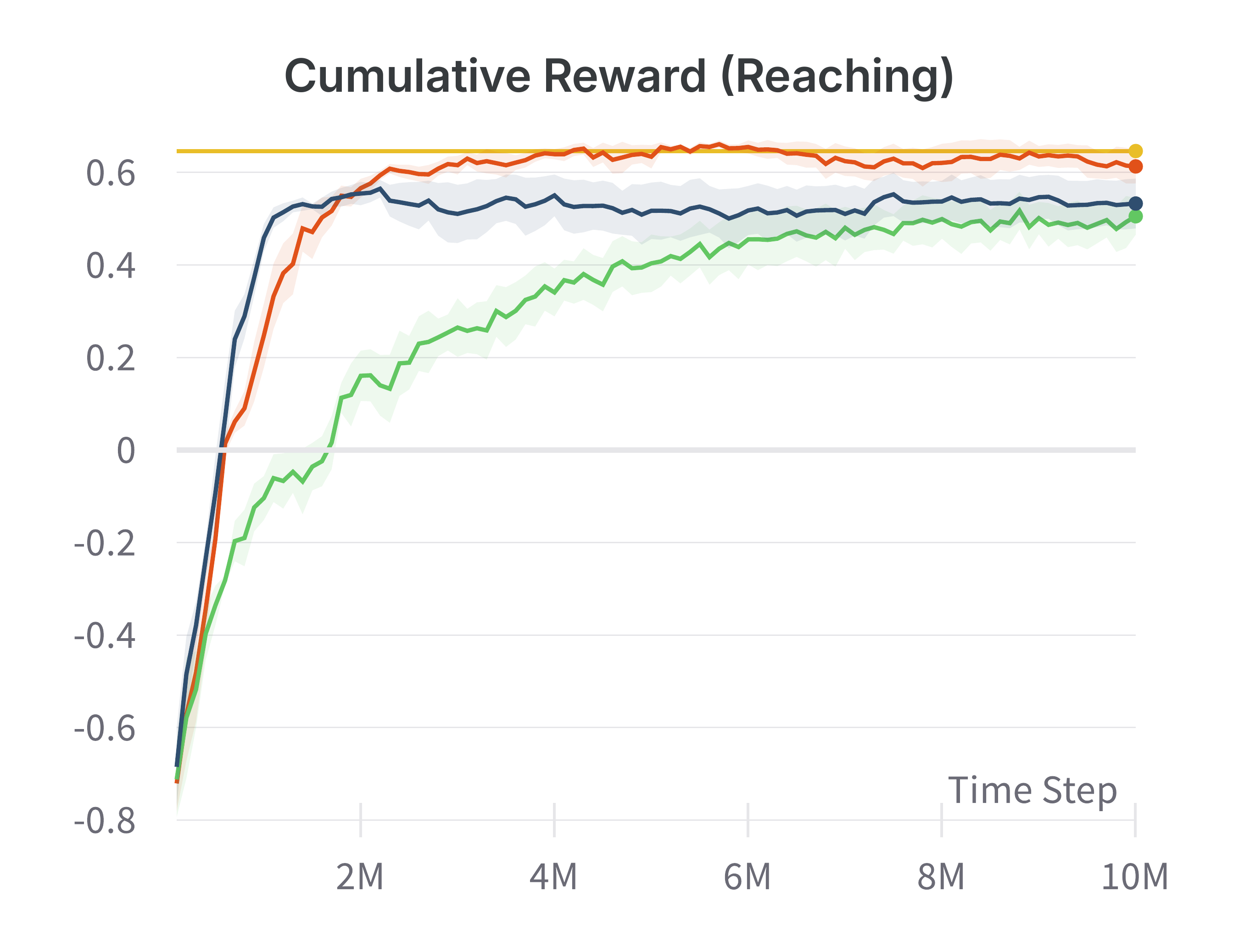}%
	\includegraphics[width=0.33\columnwidth]{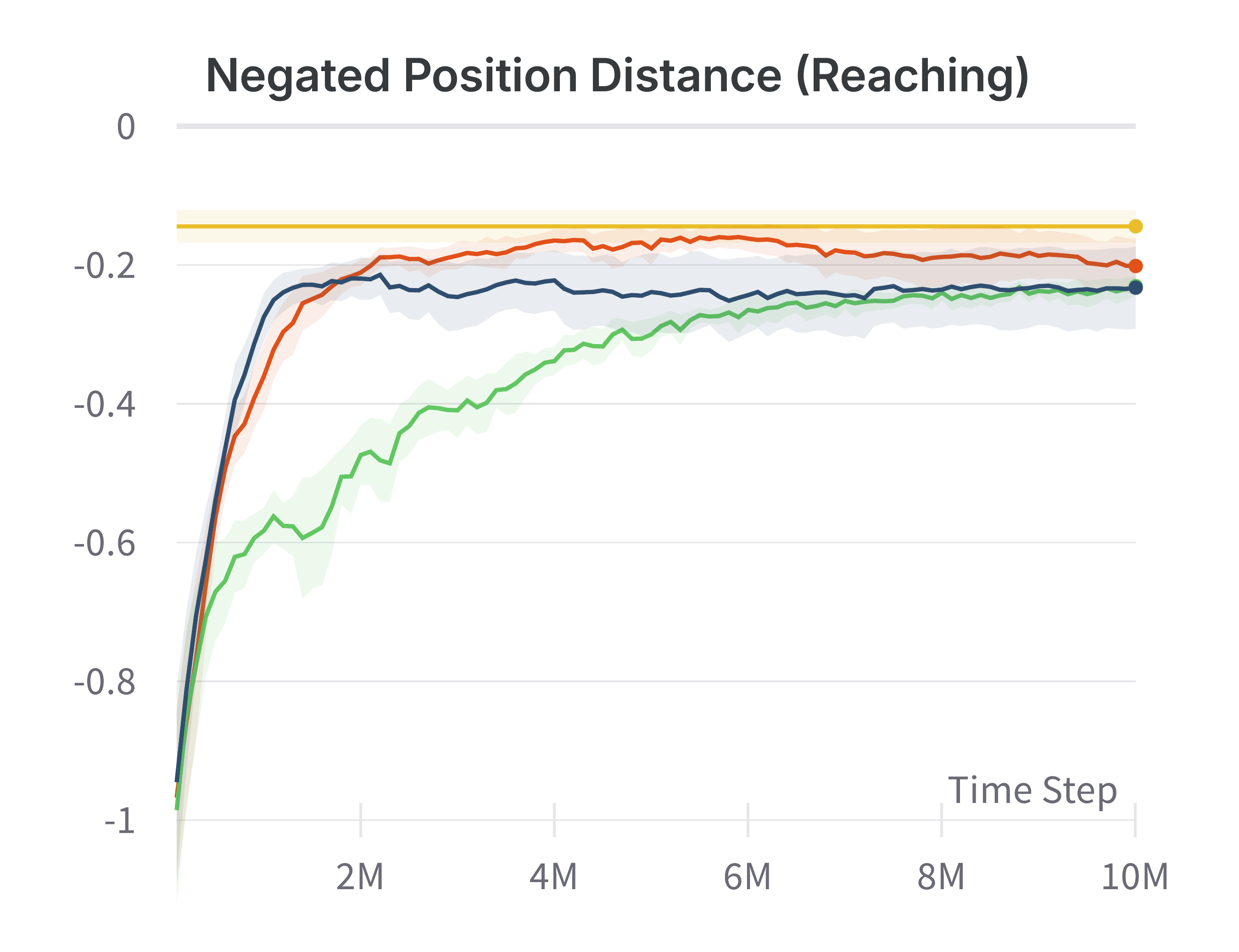}%
	\includegraphics[width=0.33\columnwidth]{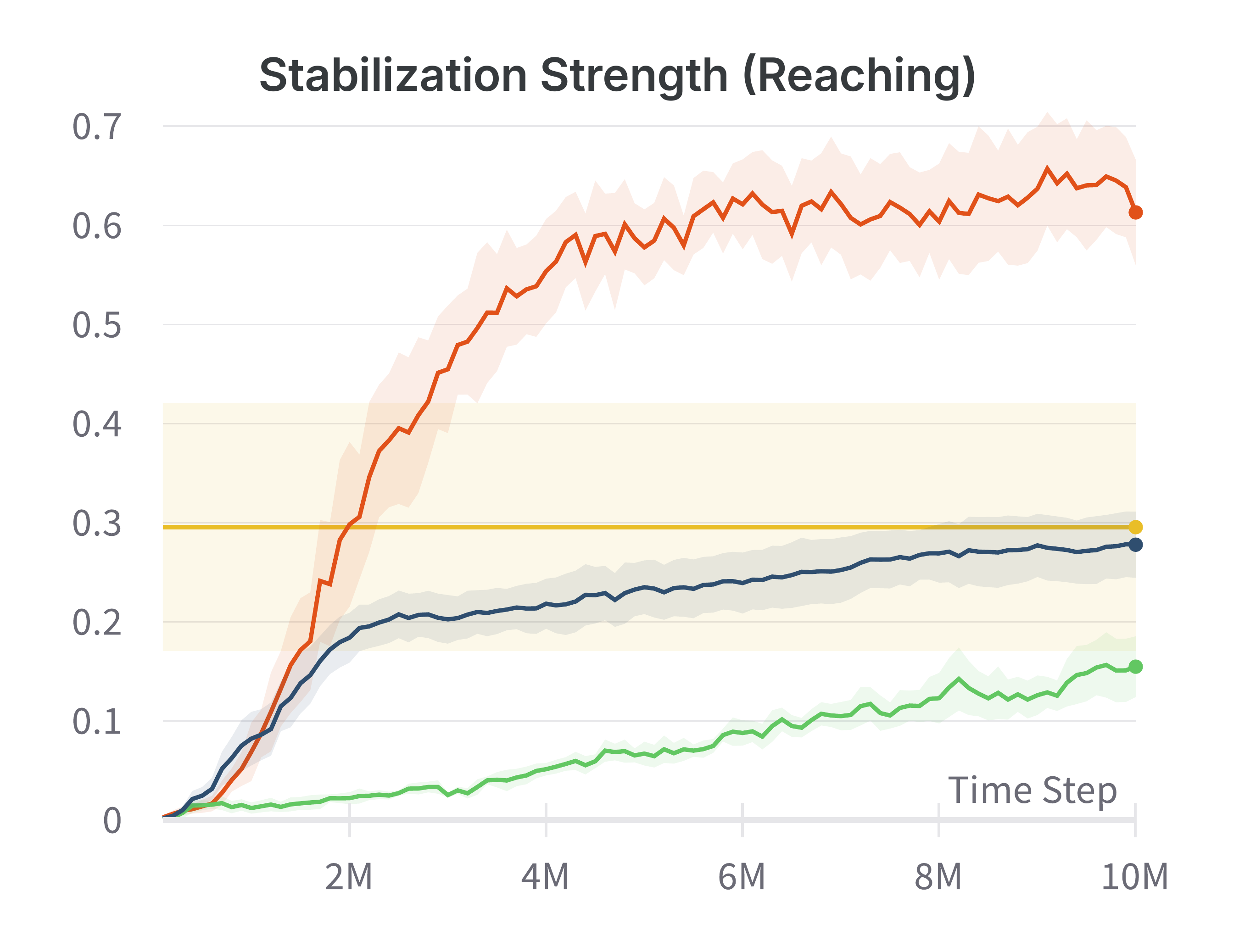}\\%
	\includegraphics[width=0.33\columnwidth]{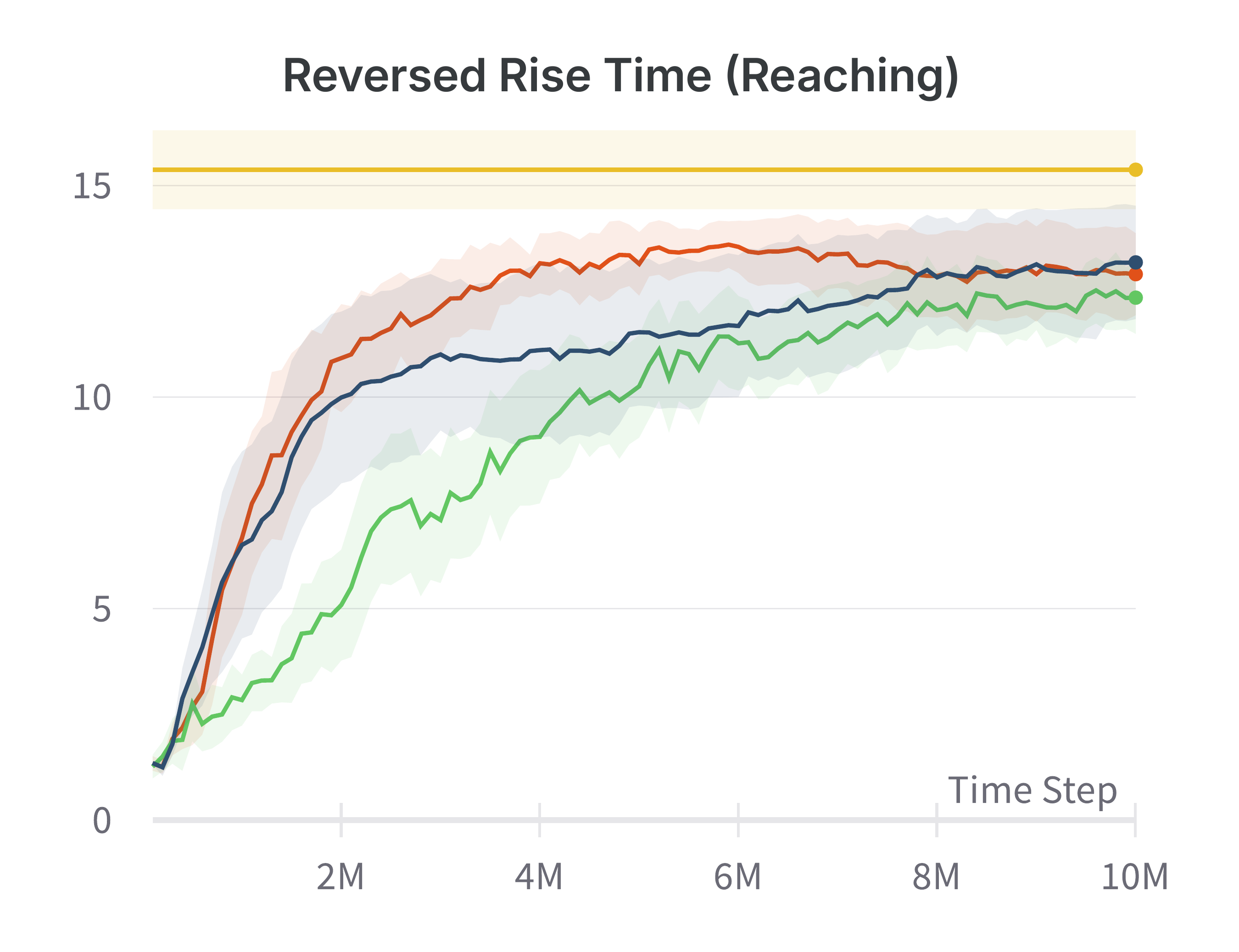}%
	\raisebox{6pt}{\includegraphics[width=0.33\columnwidth]{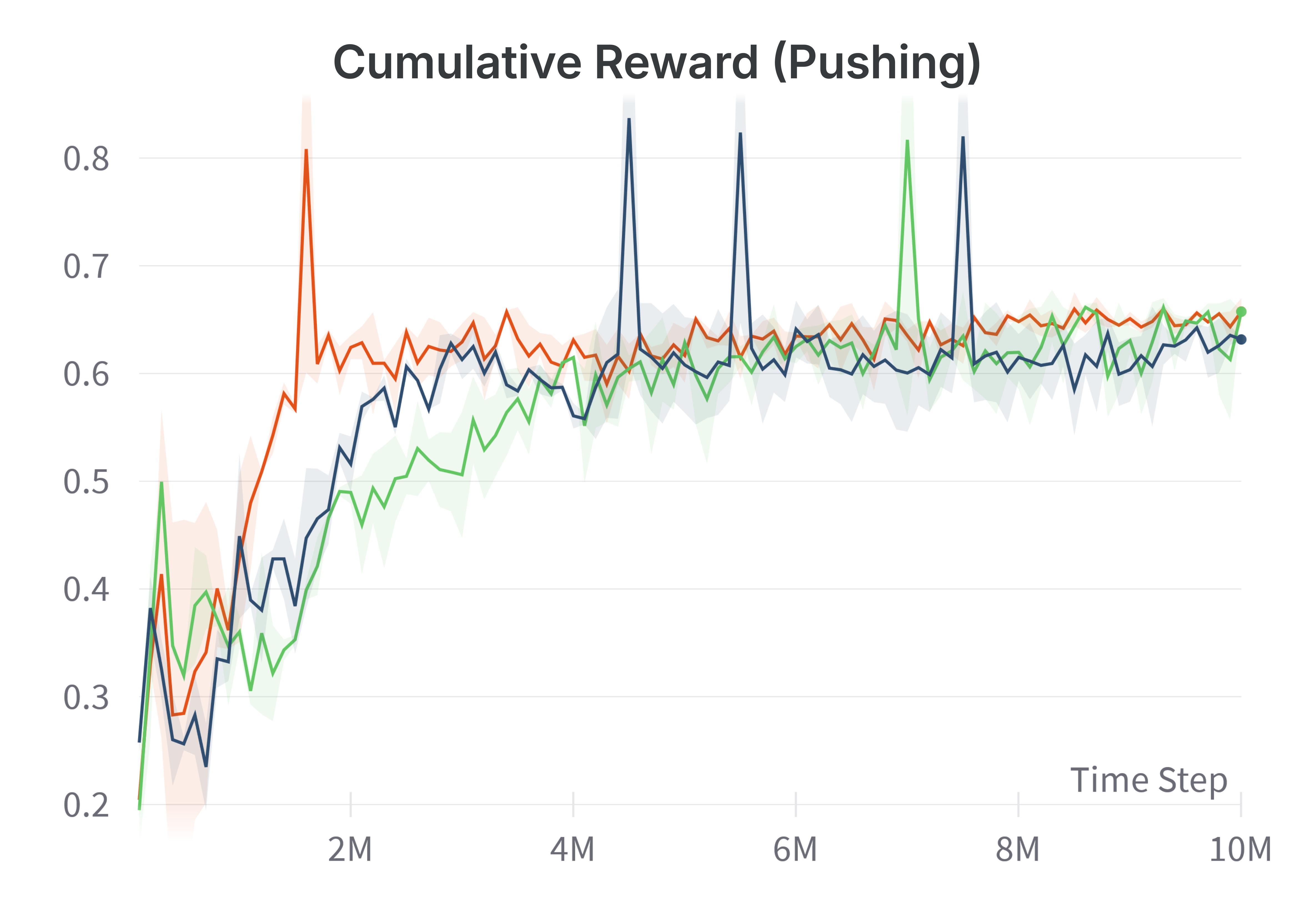}}%
	\raisebox{6pt}{\includegraphics[width=0.33\columnwidth]{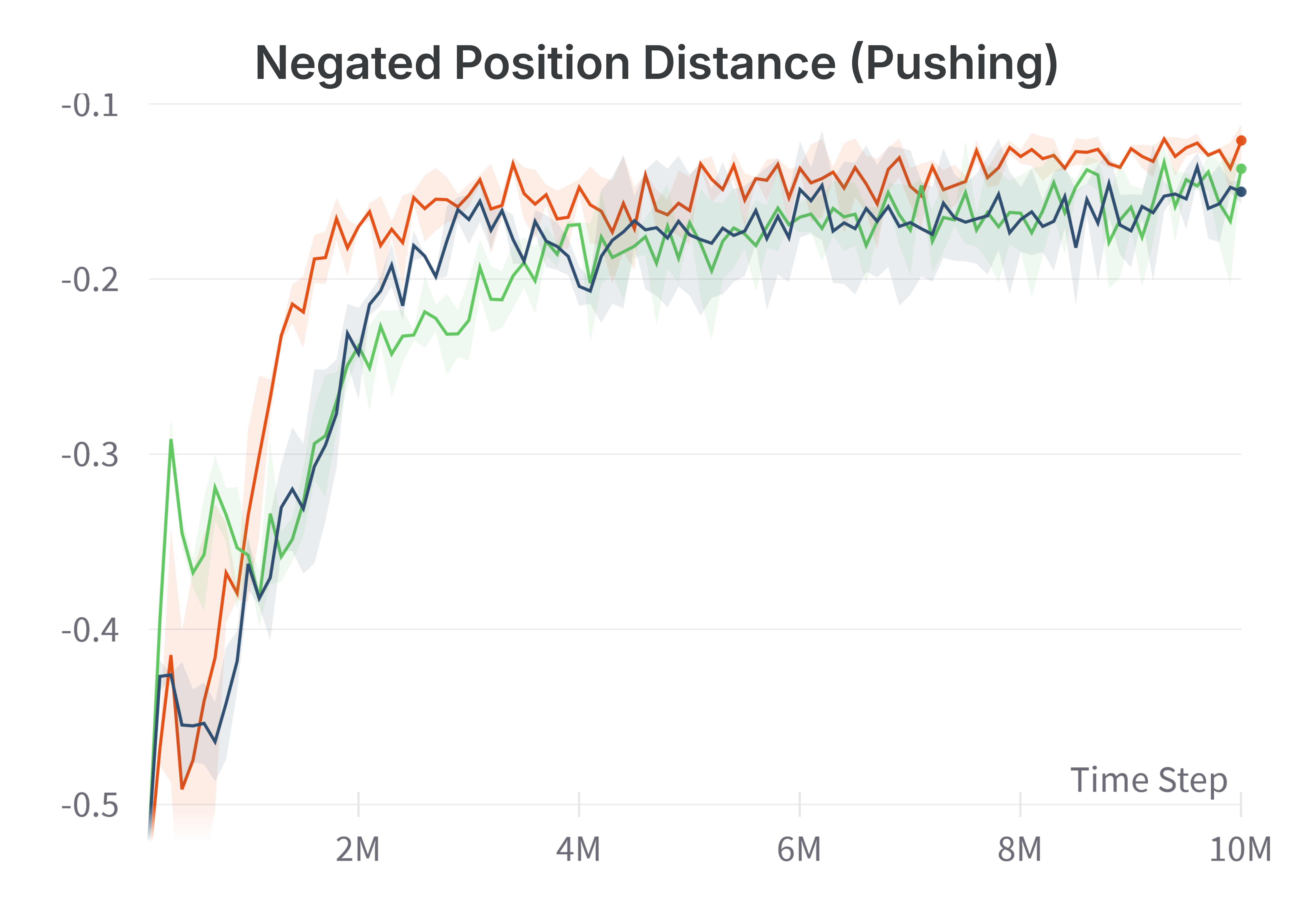}}\\%
	\vspace{-13pt}
    \makebox[\columnwidth][r]{%
	   \includegraphics[width=0.35\columnwidth]{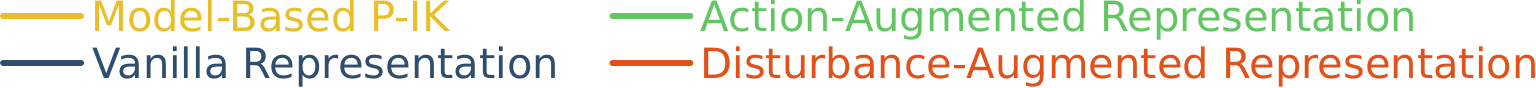}%
    }
	\vspace{-20pt}
	\caption{Evolution of test set cumulative reward and metrics during robust agent training via the three representations in the \textit{reaching} and \textit{pushing} experiments. Results are averaged across the randomization set $\Omega_0$. Shaded areas denote the standard error of the population mean.}%
	\label{fig:repr_comp_simpl}%
	\vspace{-5pt}
\end{figure}

Based on these findings, immediate rewarding is the proper choice for policy training in stochastically delayed simulated environments. For the main analysis, we trained the robust models with immediate rewarding within the budget of $K=10^7$ training time steps. To incorporate delay and disturbance randomization with a minimal yet effective setup for sim2real transfer, we used the reference set $\Omega_0 = \Theta \times \Phi_2$ for \textit{reaching} and $\Omega_0 = \Theta \times \Phi_3$ for \textit{pushing}. Bandwidth randomization was omitted as it proved unnecessary for transfer, and step disturbances in \textit{reaching} were chosen to mirror the unknown center-of-mass shifts used in \textit{pushing}. \Cref{fig:repr_comp_simpl} (top part) illustrates the cumulative reward and the three metrics for the models trained for \textit{reaching}, revealing that disturbance-augmented models consistently achieve higher cumulative reward, and best stabilization strength reaching twice the baseline. Action-augmented models exhibit a much lower convergence rate and reduced performance over all the metrics, despite having the same observation dimensionality as disturbance-augmented ones. This indicates that simply including past actions is insufficient without reliable disturbance estimates.

\begin{figure}[t]%
	\centering%
	\includegraphics[width=0.33\columnwidth]{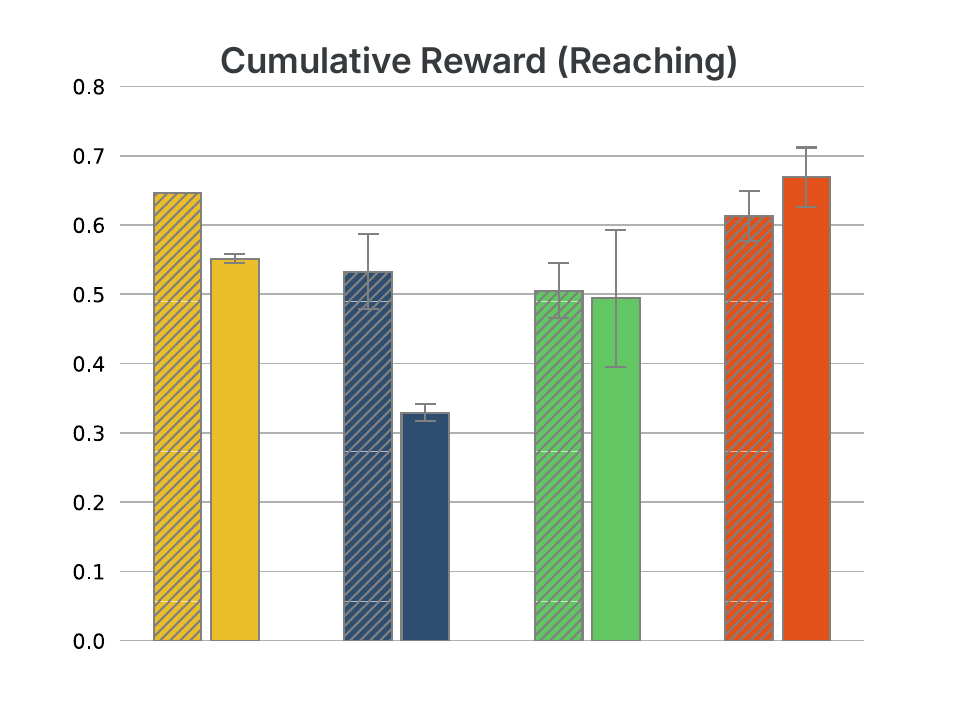}%
	\includegraphics[width=0.33\columnwidth]{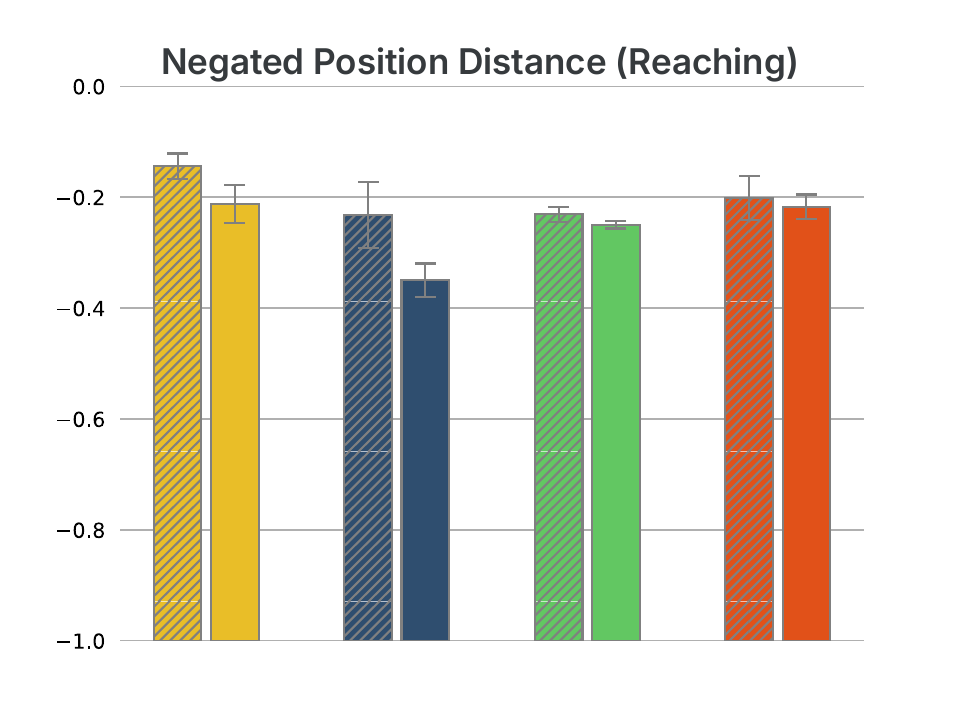}%
	\includegraphics[width=0.33\columnwidth]{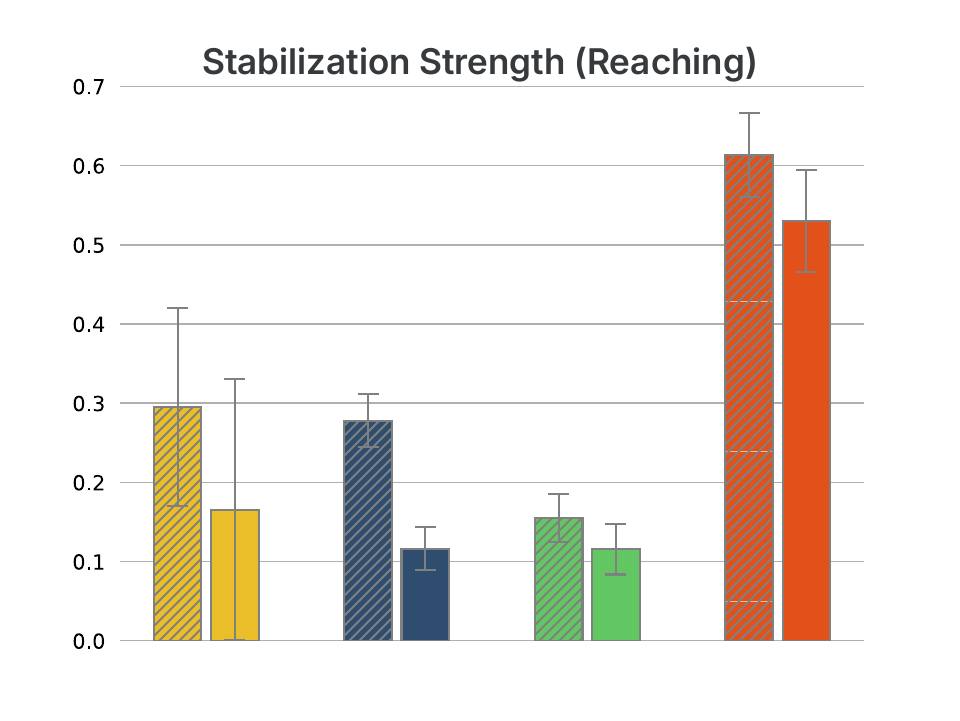}\\%
	\vspace{-5pt}
	\includegraphics[width=0.33\columnwidth]{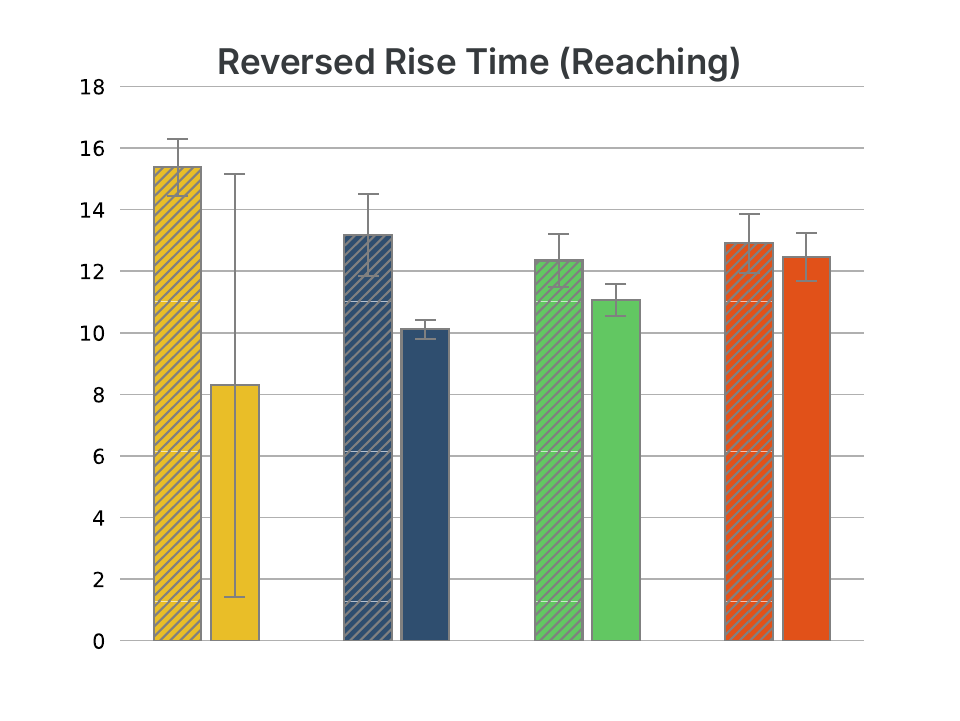}%
	\includegraphics[width=0.33\columnwidth]{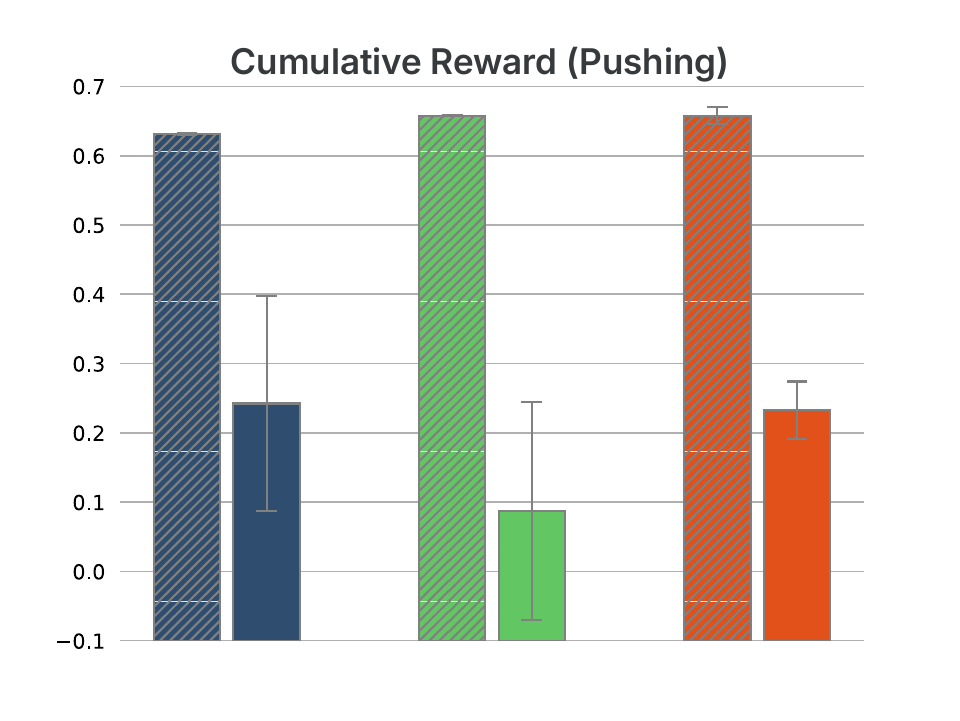}%
	\includegraphics[width=0.33\columnwidth]{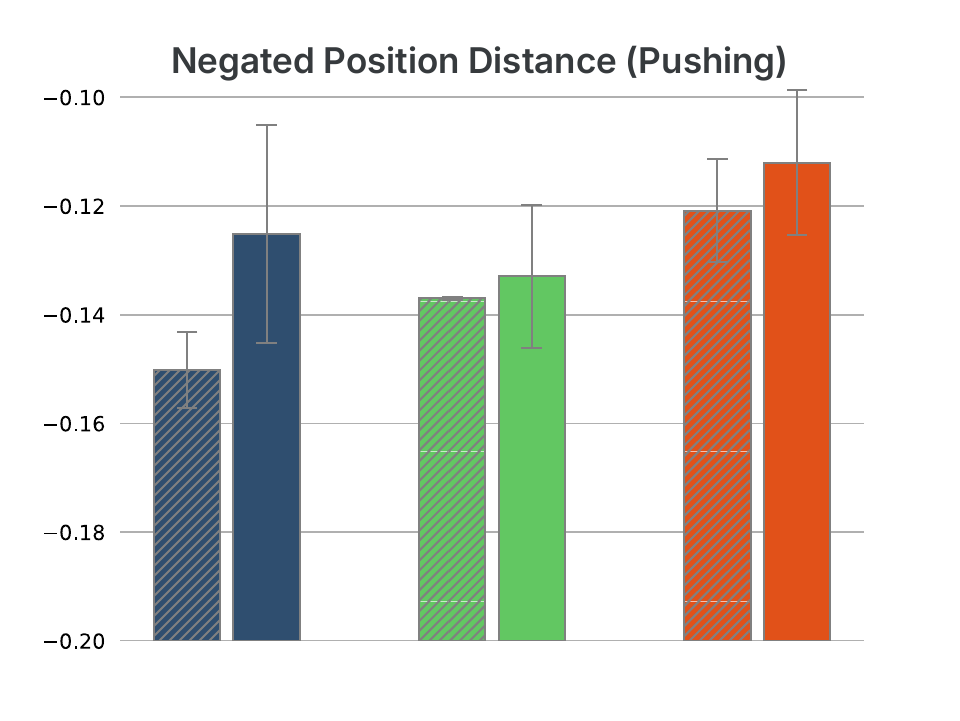}\\%
	\vspace{-10pt}
    \makebox[\columnwidth][r]{%
	   \includegraphics[width=0.4\columnwidth]{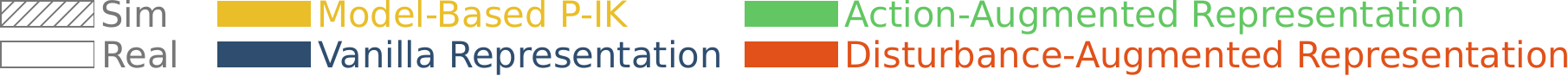}%
    }
	\vspace{-20pt}
	\caption{Comparison between the performance of the final models trained with each representation in the \textit{reaching} and \textit{pushing} experiments. The models are evaluated on the same disturbed episodes both in the simulation and the real world. Results are averaged across the randomization set $\Omega_0$. Error bars denote the standard error of the population mean.}%
	\label{fig:repr_comp_simpl_real}%
	\vspace{-5pt}
\end{figure}

From the metrics considered in \textit{reaching}, stabilization strength, defined as the reduction in local tracking error, is the most informative metric for evaluating disturbance rejection, and it is less affected by the distance to the test targets. As shown in (\Cref{fig:repr_comp_simpl_real}), disturbance-augmented models maintain significantly higher stabilization and disturbance rejection in both simulation and real-world evaluations. Vanilla models demonstrate an average oscillating motion around the targets due to their lack of disturbance estimation, while action-augmented models fail to capture sufficient disturbance features. Notably, although model-based P-IK beats all the trained agents in all the metrics in the undisturbed case, its stabilization ability degrades significantly under random disturbances, due to its lack of error-integrative action. A Welch’s t‐test on real‐world stabilization strength shows that the disturbance-augmented agent significantly outperforms the vanilla $(p=0.0040)$ and action-augmented $(p=0.0034)$ agents, while its difference with P-IK is not significant $(p=0.2371)$. The bottom part of \Cref{fig:repr_comp_simpl_real} shows that the disturbance-augmented models have superior performance as well in \textit{pushing} and can achieve higher cumulative reward and NPD both in simulation and the real system. In terms of qualitative results, these models also produce smoother stable motions with less overshoot, as demonstrated in the supplementary video material.

Although, in principle, delay-resolved action-augmented models have complete Markovian inputs, in practice, their MLP-based mid-fusion networks struggle to disentangle disturbance-relevant features end-to-end within a fixed training budget. In contrast, disturbance-augmented models converge much faster by relying on the normalized, low-variance disturbance estimates. Unlike actions $a_k$, estimated disturbances $d_k$ exhibit smaller magnitudes and a zero-centered distribution, which ease feature extraction and stabilize optimization.

\vspace{-0.5\baselineskip}
\section{CONCLUSION AND FUTURE WORK}

\noindent Systematic errors in simulating the actuation mechanism of a robot or object's contact dynamics are unavoidable without proper system identification. More often, external disturbances are conventionally assumed as additive signals to the control commands. Through this perspective, this work presented disturbance-aware reinforcement learning (DiAReL) as a suitable tool for augmenting data-driven estimation of disturbance-form uncertainties in the agent's observation space. It is empirically shown that delay-resolved disturbance augmentation helps to train agents in randomized simulations. The trained agents exhibit more robust behavior in rejecting random disturbances in simulations and when transferred to the real setup. The work helps in creating a fundamental understanding of the importance of using data-driven strategies for disturbance awareness in reinforcement learning. In future work, we will focus on extending and testing the disturbance-augmented models for designing a safe and data-driven compliant control mechanism for more complicated robotic manipulation tasks, where natural external disturbances are more common.

\vspace{-0.5\baselineskip}



\Urlmuskip=0mu plus 1mu\relax
\bibliographystyle{IEEEtran}
\bibliography{paper}

\end{document}